\documentclass[11pt,letterpaper]{article}
\usepackage{cogsys}
\usepackage[T1]{fontenc}
\usepackage{times}
\usepackage[pdftex]{graphicx} 
\usepackage[ruled,vlined]{algorithm2e}
\usepackage{algorithmicx}
\usepackage{algpseudocode}
\usepackage{amsmath}
\usepackage[hidelinks]{hyperref}  
\usepackage{natbib}
\cogsysheading{X}{20XX}{1-6}{X/20XX}{X/20XX}

\ShortHeadings{Classical AI vs. LLMs}
              {M.\ Mainali \ et \ al.}

\begin{document} 

\title{Classical AI vs. LLMs for Decision-Maker Alignment in Health Insurance Choices}
 
\author{Mallika Mainali}{mm5579@drexel.edu}
\author{Harsha Sureshbabu}{hs875@drexel.edu}
\author{Anik Sen}{as5867@drexel.edu}
\author{Christopher B Rauch}{cr625@drexel.edu}
\address{Information Science, Drexel University, 
         Philadelphia, PA 19104 USA}
\author{Noah D Reifsnyder}{Noah.Reifsnyder@parallaxresearch.org}
\address{Parallax Advanced Research, 
         4035 Colonel Glenn Hwy, Beavercreek, OH 45431 USA}
\author{John Meyer}{john.meyer@knexusresearch.com}
\author{JT Turner}{jt.turner@knexusresearch.com}
\author{Michael W. Floyd}{michael.floyd@knexusresearch.com}
\address{Knexus Research, 
          174 Waterfront Street, Suite 310, National Harbor, Oxon Hill, MD 20745 USA}
\author{Matthew Molineaux}{Matthew.Molineaux@parallaxresearch.org}
\address{Parallax Advanced Research, 
         4035 Colonel Glenn Hwy, Beavercreek, OH 45431 USA}
\author{Rosina O. Weber}{rw37@drexel.edu}
\address{Information Science \& Computer Science, Drexel University, 
         Philadelphia, PA 19104 USA}
\vskip 0.2in
 
\begin{abstract}As algorithmic decision-makers are increasingly applied to high-stakes domains, AI alignment research has evolved from a focus on universal value alignment to context-specific approaches that account for decision-maker attributes. Prior work on Decision-Maker Alignment (DMA) has explored two primary strategies: (1) classical AI methods integrating case-based reasoning, Bayesian reasoning, and naturalistic decision-making, and (2) large language model (LLM)-based methods leveraging prompt engineering. While both approaches have shown promise in limited domains such as medical triage, their generalizability to novel contexts remains underexplored. In this work, we implement a prior classical AI model and develop an LLM-based algorithmic decision-maker evaluated using a large reasoning model (GPT-5) and a non-reasoning model (GPT-4) with weighted self-consistency under a zero-shot prompting framework, as proposed in recent literature. We evaluate both approaches on a health insurance decision-making dataset annotated for three target decision-makers with varying levels of risk tolerance (0.0, 0.5, 1.0). In the experiments reported herein, classical AI and LLM-based models achieved comparable alignment with attribute-based targets, with classical AI exhibiting slightly better alignment for a moderate risk profile. The dataset and open-source implementation are publicly available at: \href{https://github.com/TeX-Base/ClassicalAIvsLLMsforDMAlignment}{https://github.com/TeX-Base/ClassicalAIvsLLMs} and \href{https://github.com/Parallax-Advanced-Research/ITM/tree/feature_insurance}{https://github.com/Parallax-Advanced-Research/ITM}.

\end{abstract}

\section{Introduction} 
The advancement of large language models (LLMs) represents a new approach to building algorithmic decision-makers (DMs) in high-stakes domains \citep{paper-liu-etal, paper-hu-etal, paper-guan-etal}. Unlike classical AI systems that rely on explicit rules and optimization procedures, LLM-based DMs draw on contextual reasoning and linguistic inference to approximate human judgment \citep{article-ravichandran-etal, article-chen-etal}, offering an alternative for algorithmic decision-making. This raises a critical question: what advantages and limitations arise from each approach, particularly in their capacity to align with human decision-making attributes? In this work, we investigate how classical AI and LLM-based DMs can be designed to achieve decision-maker alignment (DMA), which occurs when algorithmic outputs reflect the reasoning processes of human decision-makers who operate under uncertainty, pressure, and limited resources \citep{paper-sen-etal}. Under such conditions, where there is often no single correct answer, humans rely on cognitive attributes such as risk tolerance, cognitive reflection, and biases to guide their choices \citep{article-mainali-weber}. Designing algorithmic DMs that account for this variability requires methods that align not only with contextual factors but also with the diverse cognitive profiles that characterize human decision-making \citep{book-christian}.
%

Recent alignment research has largely focused on enforcing fixed ethical principles, such as honesty, helpfulness, and harmlessness, through reinforcement learning from human feedback (RLHF) \citep{article-ouyang-etal, article-bai-etal}. However, this approach does not necessarily translate to improved decision-making in dynamic, high-stakes contexts, as its rigidity limits adaptability to evolving, context-dependent scenarios \citep{article-fox}. This limitation becomes particularly salient when values diverge across individuals, influenced by situational context, personal preferences, and social trade-offs. In such cases, algorithmic decision-makers must integrate moral commonsense reasoning and mediate conflicting preferences, which extend beyond the scope of universal alignment alone \citep{article-jiang-etal, paper-sorensen-etal}. Designing algorithmic DMs that are alignable with these nuanced attributes remains a key challenge, demanding context-specific alignment that reflects how humans actually reason and decide, rather than how they are expected to choose optimally. Building on the notion of context-specific alignment,  Hu et al. \citeyearpar{paper-hu-etal} demonstrated that LLMs could be leveraged to align algorithmic decisions with cognitive attributes through zero-shot prompting and weighted self-consistency, potentially capturing nuanced human reasoning in dynamic, high-stakes contexts. 

Classical AI approaches to decision-maker alignment employ structured reasoning techniques to emulate human decision-making under uncertainty, limited resources, and time pressure \citep{paper-molineaux-etal}. Methods such as case-based reasoning, Bayesian inference, and naturalistic decision-making are employed to model how decision-makers utilize cognitive attributes when navigating complex choices \citep{paper-molineaux-etal}. For example, Molineaux et al. \citeyearpar{paper-molineaux-etal} proposed a case-based reasoning framework for combat medical triage, aligning the system outputs with decision-maker attributes using labeled prior cases. In this work, we extend a classical AI-based algorithmic decision-maker (DM) to a health insurance domain and implement an LLM-based algorithmic DM using Hu et al.’s methodology, adapting zero-shot prompting with weighted self-consistency on the same health insurance dataset and decision-maker targets defined in our experiment. This setup allows a direct comparison of how each paradigm aligns with human cognitive attributes under consistent experimental conditions.
Although classical AI approaches to decision-maker alignment have been evaluated on the health insurance domain \citep{paper-sen-etal}, LLM-based methods such as Hu et al. \citeyearpar{paper-hu-etal} have not. Moreover, Hu et al.’s implementation poses practical challenges for direct evaluation in our context, including reliance on older models (LLaMA 2) with smaller context windows and lower reasoning performance compared to newer models, the need to download large model weights, and integration within a federated system that depends on external servers. These limitations make it difficult to apply the original implementation to our dataset and decision-maker targets. Our goal is to fill this gap by re-implementing the LLM-based methodology and evaluating both classical AI and LLM-based DMs on the same health insurance dataset with the same decision-making attribute, \textit{risk tolerance}, designed specifically for our experiments. 

For the LLM-based algorithmic DM, we develop a decision-maker using Hu et al.’s methodology to ensure reproducibility and compatibility with our experimental setup. We evaluate two models: GPT-5, a reasoning model, and GPT-4, a non-reasoning model, both using weighted self-consistency with zero-shot prompting. For the classical AI algorithmic DM, we re-implement the algorithm proposed by Molineaux et al. \citeyearpar{paper-molineaux-etal}, and evaluate both DMs on the same target decision-makers. This design enables a direct comparison of alignment accuracy across individuals with varying levels of risk tolerance. Our contributions are twofold: (1) we introduce a LLM-based DM tailored to risk-sensitive decision-making, fully self-contained without reliance on federated systems or external servers, and (2) we provide a systematic comparative analysis of classical AI and LLM-based DMs, offering insights into the design of adaptive, cognitively grounded algorithmic decision-makers.

The remainder of this paper is organized as follows. Section 2 reviews related work on algorithmic decision-makers and alignment strategies. Section 3 details our methodology, including the construction of the classical AI baseline, the LLM-based decision-maker, and the evaluation metric applied to the health insurance dataset. Section 4 reports results, highlighting the comparative strengths and trade-offs of each approach. Section 5 concludes with our key findings and the directions for future research.

\section{Background} 
In this section, we first review research on decision-maker alignment, highlighting foundational work in value alignment and recent developments in pluralistic alignment. We then discuss both classical approaches and emerging methods that leverage LLMs to capture user- and domain-specific variability. Finally, we provide an overview of the dataset used in our experimental study.

\subsection{Decision-Maker Alignment}

Human decision-making is guided by internal mental models that shape how individuals perceive alternatives, evaluate risks, and select courses of action \citep{article-chermack}. A central finding across cognitive science and psychology is that choices often diverge from the assumptions of rational choice theory, particularly in uncertain or high-stakes environments \citep{article-tversky-kahneman}. In such contexts, decision makers are frequently forced to select suboptimal options due to environmental or cognitive limitations \citep{article-simon}. 

Decision-Maker Alignment (DMA) refers to the design of algorithms that align with the values and decision-making tendencies of the individual decision-makers. In scenarios where optimal decisions are unavailable, DMA focuses on modeling the variability in human choices by ensuring that the model's decisions align with the attributes that guide each decision-maker \citep{paper-molineaux-etal, paper-sen-etal}. By explicitly considering different levels of decision-maker attributes, DMA provides a framework for creating AI systems that can adapt to differing perspectives and reasoning patterns.

\subsection{Value Alignment vs. Pluralistic Alignment}
Value alignment approaches aim to align algorithmic decision-makers with a fixed set of normative values. Standard methods often employ reinforcement learning from human feedback (RLHF), where a reward model is trained on human preference data to shape outputs in accordance with broad values such as honesty, harmlessness, and helpfulness \citep{article-ouyang-etal}. While effective at providing coarse reward signals, these methods are limited to static notions of human intent. To support broader alignability, benchmark datasets like ETHICS enable algorithmic DMs to align with values such as justice, well-being, virtues, commonsense morality, and duties \citep{paper-hendrycks-etal}. Shen et al. \citeyearpar{article-shen-etal} further developed ValueCompass to evaluate alignment across real-world scenarios using 56 value statements from Schwartz’s Theory of Basic Values, highlighting the variability of values and limitations of static strategies \citep{article-shen-etal}.

These limitations have motivated more flexible alignment methods. Domain-aware alignment incorporates ethical principles along with domain-specific knowledge \citep{paper-shetty-etal, paper-rauch-etal}, while pluralistic alignment explicitly models cognitive and behavioral variation across users \citep{paper-sorensen-etal}. Such approaches capture human reasoning variability by aligning AI systems with personalized mental models, user preferences, and decision styles \citep{paper-chen-etal, paper-wu-etal}. Wu et al. \citeyearpar{paper-wu-etal} proposed a dynamic method in which LLMs infer and adapt to user preferences through multi-turn interactions, while Guan et al. \citeyearpar{paper-guan-etal} introduced a unified framework integrating preference memory, feedback-driven refinement, and individualized generation. Together, these works illustrate a shift from static value alignment toward pluralistic, context-dependent approaches that better reflect the diversity of human values and decision-making attributes, particularly for high-stakes algorithmic DMs.

\subsection{Classical AI Approaches to Decision-Maker Alignment}
While broader types of alignment approaches have been experimented with using expert knowledge, utility theory, and human-in-the-loop designs \citep{article-newell-simon, article-zhi-etal}, they do not address the alignment we focus on here, which is user-specific, context-specific alignment with decision-making. To date, the only work exploring this practical and individual alignment using classical AI methods is Molineaux et al. \citeyearpar{paper-molineaux-etal}, who developed an algorithmic DM combining case-based reasoning, Monte Carlo simulation, Bayesian diagnosis, and naturalistic decision-making and demonstrated improved alignment in simulated military triage scenarios.

\subsection{LLM-Based Approaches to Decision-Maker Alignment}
Recent advances in LLMs have expanded the focus of alignment research from explicit knowledge engineering to scalable alignment methods via preference modeling, prompt engineering, and human feedback \citep{paper-hu-etal}. Unlike classical expert systems, which directly encode domain rules, LLMs acquire broad decision-relevant knowledge through large-scale pretraining and are then steered toward alignment with human values using techniques such as reinforcement learning from human feedback (RLHF) \citep{paper-christiano-etal, article-ouyang-etal}. RLHF emphasizes learning reward functions from human demonstrations and preferences, enabling models to generate outputs that are more consistent with human preferences. Using reinforcement learning, methods such as constitutional AI introduce alignment through explicit normative principles and self-critique techniques \citep{article-bai-etal}, while iterative preference optimization refines model behavior through ongoing user interaction \citep{article-zeng-etal}. 

Another line of research involves steerability and controllability frameworks, where alignment is addressed through prompts, fine-tuning, and policy constraints \citep{article-ganguli-etal}. This includes persona-based alignment, in which prompts describing specific personas are used to steer LLMs toward responses reflective of different demographic or ideological groups \citep{paper-santurkar-etal, paper-hwang-etal}. Recent studies have expanded persona-based alignment methods to incorporate domain-specific attributes for high-stakes scenarios. As demonstrated by Hu et al. \citeyearpar{paper-hu-etal}, incorporating decision-maker attribute information directly into prompts, combined with a self-consistency module to weigh positive and negative samples, significantly improved alignment of LLM-based algorithmic DMs. Building on their methodology, we develop a LLM-based DM and apply it to a new domain, enabling a systematic comparison of both classical AI and LLM-based approaches.

\begin{figure}[t]
\vskip 0.05in
\begin{center}
\includegraphics[width=0.85\linewidth]{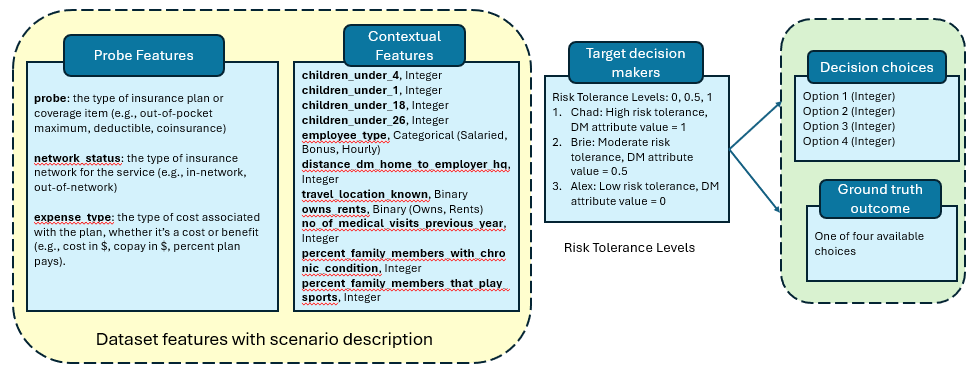}
\caption{Schematic overview of the dataset structure with example probes, contextual attributes of the decision-maker, a target decision maker attribute, \textit{risk tolerance}, and four available choices. The ground truth indicates the most aligned option.} 
\label{data-schema}
\end{center}
\vskip -0.2in
\end{figure} 
\subsection{Dataset}
One major challenge in decision-maker alignment is the lack of adequate datasets that capture context, target decision-makers, and their influences. To address this, Sen et al. \citeyearpar{paper-sen-etal} created a health insurance benchmark dataset designed to train and evaluate algorithmic decision-makers. The dataset simulates dilemmas individuals face when selecting health insurance plans under uncertainty, comprising 20 cost-related probes derived from real plan specifications (\textit{e.g.,} deductibles, out-of-pocket maximums, specialist visits), each paired with alternative plan options in integer dollar amounts.

Decision-maker behavior is characterized by two cognitive attributes: risk tolerance and choice, each labeled as high or low, yielding four possible combinations. Risk tolerance captures willingness to accept risk (\textit{e.g.,} selecting lower premiums with higher deductibles), while choice reflects preferences for the number of alternatives for services. In addition to decision-making attributes, the dataset incorporates contextual features such as family composition, medical history, employment type, and lifestyle factors (\textit{e.g.}, sports participation, chronic conditions) to describe the scenarios. Each instance includes four plan options, a labeled attribute, and the ground-truth choice aligned with the target decision-maker. 

\section{Methodology}
In this section, we first outline the classical alignment approach used in our work. We then introduce our LLM-based alignment method, followed by a description of the evaluation framework used to assess alignment in the algorithmic decision-maker responses across both approaches. Our hypothesis is that the two approaches will achieve similar levels of alignment. We then present experimental results comparing the alignment achieved by each approach across different target profiles.

\subsection{Dataset}
Sen et al. \citeyearpar{paper-sen-etal} created a health insurance dataset for four target decision-maker types. For our experiments, we aimed to capture all variations of target attribute levels and explicitly model the effect of risk tolerance on individual decisions. Specifically, for each probe and each possible plan option, we sought to determine the level of risk tolerance that would be associated with a decision-maker selecting that option.

For all probes and decisions, the original dataset defined four attribute combinations, \textit{low–low, low–high, high–low, and high–high}, which corresponded to 0.01, 0.34, 0.67, and 1.0, respectively. Using these four attribute values, we created a regression neural network that computed risk tolerance levels for all probes across all alternative choices. Figure~\ref{data-schema} presents a schematic overview of the dataset, highlighting example probes and contextual features.


The original dataset contained 32,000 probes, with 24,000 associated with risk tolerance and 8,000 associated with choice. Because choice probes were less explicit, we excluded them from our experiments. This exclusion introduced duplicates, as the only distinguishing feature for many probes was the choice attribute. After removing duplicates, 17,400 distinct probes remained. From this set, we randomly sampled 5,000 probes for training and 1,000 probes for testing. To evaluate alignment, we instantiated three synthetic targets, \textit{Chad}, \textit{Brie}, and \textit{Alex}, with risk tolerance levels of 1.0 (highly risk-tolerant), 0.5 (moderately risk-tolerant), and 0.0 (highly risk-averse), respectively. Because ground-truth choices differ by target, the full set of probes was administered three times, once per target. This design allowed us to assess model performance across distinct risk preferences and provided a controlled framework to analyze how decision-making behavior varies with target risk tolerance.

\subsection{Classical algorithmic DM}
In this work, we employ the model proposed by Molineaux et al. \citeyearpar{paper-molineaux-etal} as a classical AI algorithmic DM. This model uses scenario features formalized through four components: probes $p \in P$, representing decision environments, decisions $y \in Y$, which include all the possible actions, contextual categories $CC$, which describe features of the scenario, and decision-maker attributes $u \in U$, which include the decision-making values influencing the decision process. The model functions in two phases: during offline training, it builds a case base by generating candidate decisions for each probe, evaluating them with Monte Carlo simulations, Bayesian reasoning, and heuristic-based rationalizers, and linking these decisions to target decision-makers. In the online stage, when new probe and target information are provided, it creates and analyzes the possible decisions, searches for similar cases in the case base using learned similarity weights, and selects the action that most closely aligns with the specified profile. This process enables the model to approximate diverse human decision-making styles in high-stakes environments where no single optimal decision exists. The high-level pseudocode for this algorithm is presented in Algorithm ~\ref{alg:TAD}.

\begin{algorithm}[H]
\caption{Classical algorithmic DM for Aligned Decision-Making}
\label{alg:TAD}
\KwIn{Probe $p$ with context $x$, target DM $dm$, target DM attribute value $dma$}
\KwOut{Aligned Decision consistent with the target decision-maker}

\BlankLine
\textbf{Offline Training Subsystem:} \\
\ForEach{$p$ with known decision}{
    PossibleDecisions $\leftarrow$ DecisionSpaceElaboration($p$) \tcp{Get possible decisions}
    decisionAnalytics $\leftarrow$ DecisionAnalysis(PossibleDecisions) \\
    targetInfo $\leftarrow$ targetDetectionFunction($p$, decision) \\
    case $\leftarrow$ ($p$, decision, decisionAnalytics, targetInfo) \\
    Add case to AlignmentCaseBase 
}
weights $\leftarrow$ AlignmentWeightLearning(AlignmentCaseBase)

\BlankLine
\textbf{Online Decision-Making Subsystem:} \\
\ForEach{new probe}{
    PossibleDecisions $\leftarrow$ DecisionSpaceElaboration($p$) \\
    decisionAnalytics $\leftarrow$ DecisionAnalysis(PossibleDecisions) \\
    \ForEach{decision $d \in$ PossibleDecisions}{
        case$_d \leftarrow$ ($p$, $d$, decisionAnalytics, TargetInfo) \\
        neighbors $\leftarrow$ RetrieveNearestNeighbors(case$_d$, AlignmentCaseBase, weights) \\
        estimatedTarget $\leftarrow$ WeightedAverage(neighbors) 
    }
    alignedDecision $\leftarrow \arg\min\limits_{d} \big| \text{estimatedTarget} - \text{targetInfo} \big|$ 
}
\Return{alignedDecision} 
\end{algorithm}

\subsection{LLM-Based algorithmic DM}
We use a large reasoning model, GPT-5\textsuperscript{TM} \citep{misc-openai2025-etal}, and a large language model, GPT-4\textsuperscript{TM} \citep{article-openai-etal} via the OpenAI API, configured with a temperature of 0.7 to evaluate their capacity for alignment with cognitive attributes across diverse decision-making scenarios. To create an aligned LLM-based DM, we implement the methodology introduced by Hu et al. \citeyearpar{paper-hu-etal}, originally developed for context-aligned decision-making in the medical triage domain. This framework operationalizes alignment of algorithmic DMs by associating decisions with key decision-maker attributes such as utilitarianism, fairness, and risk aversion.

\begin{algorithm}[H]
\caption{LLM-Based Algorithmic DM with weighted self-consistency as proposed by Hu et al.}
\label{alg:llm_alignment_selfconsistency}
\KwIn{LLM/LRM, Dataset $\mathcal{D}$ (context $x$, question $q$, choices $C$, target DM $dm$, target DM attribute value $dma$), target-aligned prompts $\mathcal{P}$}
\KwOut{Aligned decision consistent with the target decision-maker}

\BlankLine
\ForEach{$(x, q, C, dm, dma) \in \mathcal{D}$}{
    Initialize vote counts $V[c] \gets 0$ for each $c \in C$

    \BlankLine
    \textbf{Weighted Self-Consistency Sampling:}

    \BlankLine
    \textit{(1) Positive Sampling: Target DM}
    
    \For{$N = 1$ \KwTo $5$}{
        $prompt^{+} \gets \text{FormatPrompt}(x, q, C, \mathcal{P}[dm][dma])$ 
        $resp^{+} \gets \text{QueryLLM}(model, prompt^{+}, \text{temperature}=0.7)$ 
        $ans^{+} \gets \text{ExtractAnswer}(resp^{+})$ \\
        $V[ans^{+}] \gets V[ans^{+}] + 1$ 
    }

    \BlankLine
    \textit{(2) Negative Sampling: Inverse DMs}
    
    Determine inverse (non-target) DMs: $\text{neg\_targets} \gets \{ \text{Alex}, \text{Brie}, \text{Chad} \} \setminus \{ dm \}$ 
    Randomly choose allocation pattern: $(n_1, n_2) \gets (2,3)$ or $(3,2)$ 

    \ForEach{$(\text{invDM}, n_{\text{invDM}})$ in zip(neg\_targets, [n$_1$, n$_2$])}{
        \For{$N = 1$ \KwTo $n_{\text{invDM}}$}{
            $prompt^{-} \gets \text{FormatPrompt}(x, q, C, \mathcal{P}[\text{invDM}][dma])$ \tcp{Use inverse DM prompt}
            $resp^{-} \gets \text{QueryLLM}(model, prompt^{-}, \text{temperature}=0.7)$ 
            $ans^{-} \gets \text{ExtractAnswer}(resp^{-})$ \\
            $V[ans^{-}] \gets V[ans^{-}] - 1$ 
        }
    }

    \BlankLine
    \textbf{Final Aggregation:} \\ 
    alignedDecision $\gets \arg\max_{c \in C} V[c]$ 
    
    \Return{alignedDecision} 
}
\end{algorithm}

Each experimental scenario includes a contextual description and a decision-making question, along with all possible answer choices. We create three individual prompts, each aligned with a specific target decision-maker. For each probe, these prompts are presented separately to the three targets, Chad, Alex, and Brie, and the model selects the decision corresponding to the assigned target profile. For each scenario, the model is first given a scenario prompt outlining the structure and salient features of the task. We then apply a zero-shot prompting strategy for target-specific alignment prompts, designed to steer the model’s response toward the target attribute value.

\begin{figure}[t]
\vskip 0.05in
\begin{center}
\includegraphics[width=0.85\linewidth]{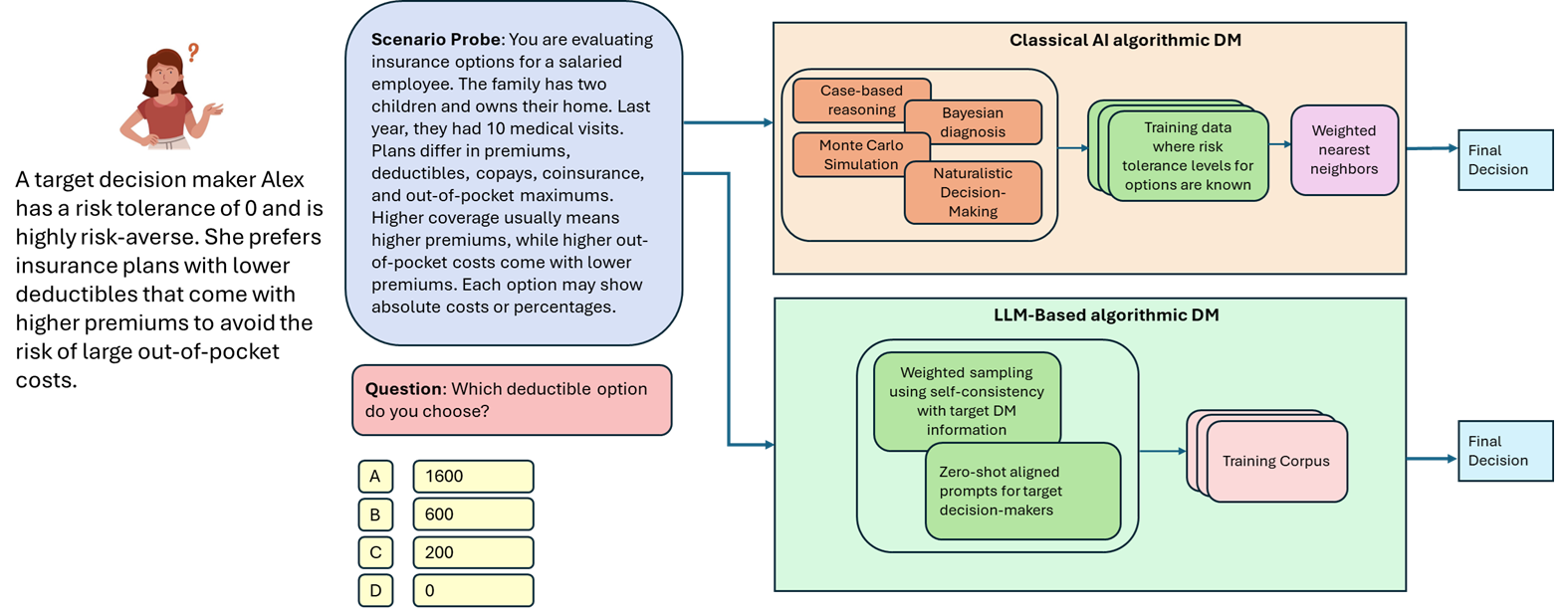}
\caption{Implementation of classical AI and LLM-based algorithmic DMs, where both models receive a scenario probe and output a final decision. The final decision must be aligned with the decision made by a target decision-maker of same risk level. The classical AI algorithmic DM relies on prior cases, while the LLM-based DM uses zero-shot prompting and self-consistency sampling for aligned decision-making.} 
\label{algorithm}
\end{center}
\vskip -0.2in
\end{figure}

Additionally, we integrate the weighted self-consistency module \citep{paper-hu-etal, article-wang-etal}, allowing the model to stabilize its decision tendencies across multiple generations while preserving context sensitivity. This process involves issuing multiple queries under both positive prompts (aligned with the target attribute) and negative prompts (aligned with the opposite target attribute), producing a diverse set of candidate responses across multiple runs. We drew N=5 samples for the positive target, and for N=5 for the negatives. Since our experiment had three targets, we split the negative samples randomly. For instance, if Brie was the positive target, we generated 5 positive samples using the prompt aligned to Brie, and randomly generated 2 samples using \textit{Alex-aligned} prompt and 3 using \textit{Chad-aligned} prompt, which were given negative weights. The final decision is determined via a weighted majority voting scheme, which emphasizes responses consistent with the target DM attribute while down-weighting outputs generated by negative prompts. This mechanism reduces stochastic variability and ensures that the generated decisions are explicitly conditioned on the intended risk tolerance levels. The high-level pseudocode for this approach is presented in Algorithm ~\ref{alg:llm_alignment_selfconsistency}. Figure~\ref{algorithm} illustrates the distinct decision-making pipelines of classical and LLM-based algorithmic DM approaches in our implementation.

\subsection{Evaluation Metric}
To evaluate model predictions, we measured alignment between model-generated actions and the ground-truth decisions of the three targets,\textit{Alex}, \textit{Brie}, and \textit{Chad}. For each target, an action was assigned a value of 1 if it matched the corresponding ground-truth decision (aligned) and 0 otherwise (misaligned).

Formally, for each scenario, let $gd$ be the ground-truth decision for a given risk tolerance value $dma$, and let $alignedDecision$ be the model’s predicted decision for the same level of risk tolerance. The alignment score is computed as:
\[
a(gd, alignedDecision, dma) =
\begin{cases}
1 & \text{if } gd = alignedDecision \\
0 & \text{otherwise}
\end{cases}
\]

We evaluate model performance using two levels of accuracy:
\begin{itemize}
    \item \textbf{Individual Target Accuracy:} Alignment is measured separately for each target (Alex, Brie, and Chad), capturing performance for different levels of risk tolerance. Formally, for a target $dm$:
    \[
        A_{\text{target}}(dm) = \frac{1}{|\mathcal{D}|} \sum_{(x,q,C,dma) \in \mathcal{D}} a(gd, alignedDecision, dma),
    \]
    where $|\mathcal{D}|$ is the total number of probes, $(x,q,C,dma) \in \mathcal{D}$ denotes each probe with its context $x$, question $q$, choice set $C$, and decision-making attribute $dma$. The function $a$ gives a score of 1 if the model's predicted action $alignedDecision$ matches the ground truth $gd$ for target attribute $dma$, and 0 otherwise.

    \item \textbf{Overall Accuracy:} To summarize performance over the entire dataset, we average across all probes and targets:
        \[
            A_{\text{total}} = \frac{1}{|\mathcal{D}| \times 3} \sum_{dma \in \{0,0.5,1\}} \sum_{(x,q,C,dma) \in \mathcal{D}} a(gd, alignedDecision, dma)
        \]
\end{itemize}

\section{Results and Discussion}

\begin{figure}[t]
\vskip 0.05in
\begin{center}
\includegraphics[width=0.80\textwidth]{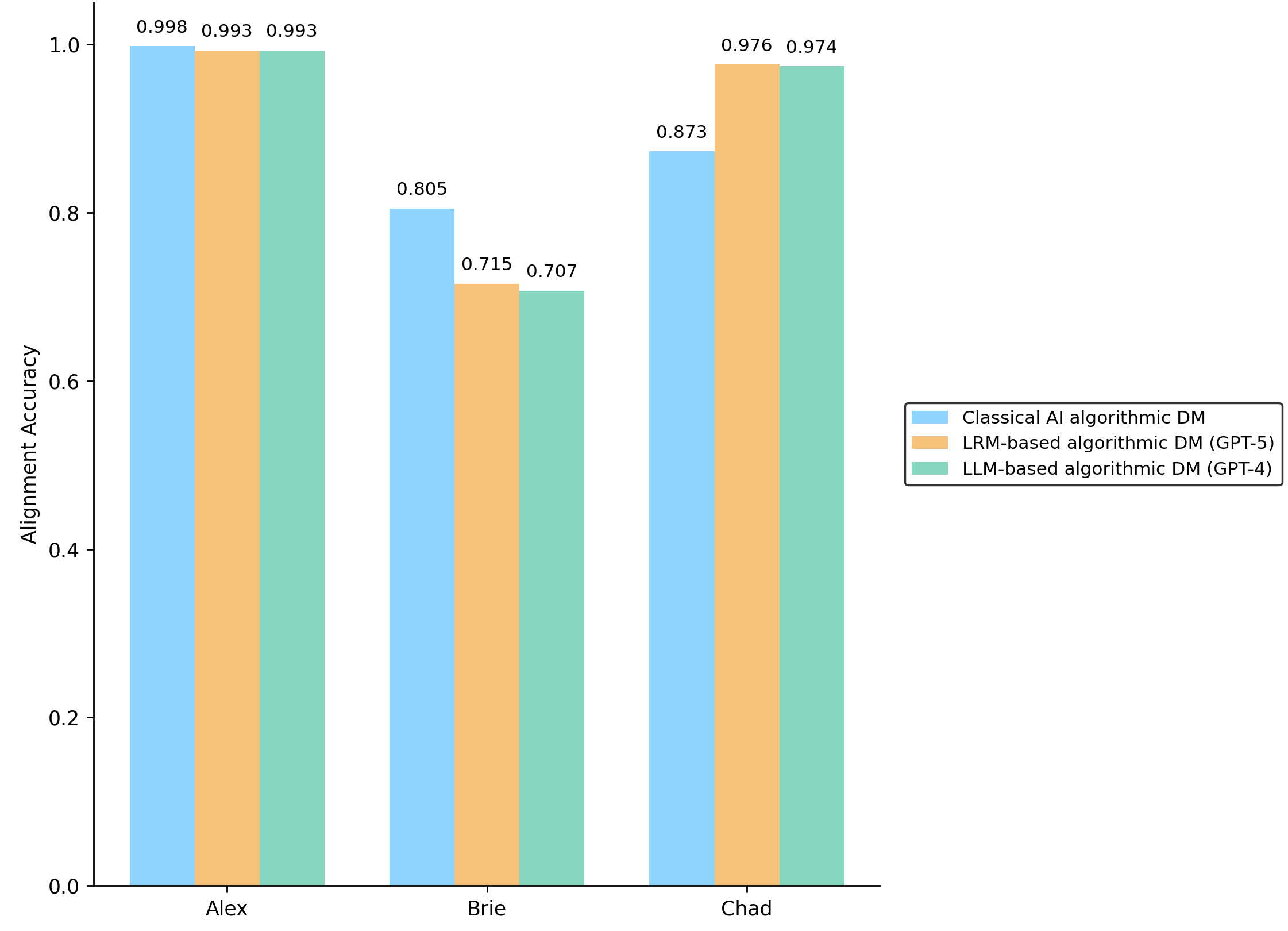}
\caption{Performance of the three models across three targets with varying risk tolerances (Alex: $0$, highly risk-averse; Brie: $0.5$, moderately risk-averse; Chad: $1.0$, risk-tolerant). Bars indicate individual target alignment accuracy, and the legend denotes the model.} 
\label{fig:all-results}
\end{center}
\vskip -0.2in
\end{figure}

In this section, we report results for both decision-making paradigms. The primary findings are summarized in Figure~\ref{fig:all-results} and Table~\ref{tab:overall-accuracy}. Across all models, alignment accuracy was very similar; however, the GPT-5-based algorithmic DM achieved the highest overall performance across the dataset.
\subsection{Comparison of Classical AI and LLM-Based Algorithmic DMs}
Our experiments demonstrated comparable overall performance between the classical AI algorithmic DM and the two LLM-based DMs, with all three models achieving nearly identical alignment accuracy, as shown in Table~\ref{tab:overall-accuracy}. Both approaches performed best for the target with the lowest level of risk tolerance, showing near-perfect alignment, suggesting that the models effectively capture the deterministic patterns associated with highly risk-averse decision-making in our dataset. Although the LLM-based algorithmic DMs achieved near-perfect alignment for the two extreme targets when using weighted self-consistency, their accuracy dropped for the target with a moderate level of risk tolerance, whereas the classical AI algorithmic DM maintained relatively good alignment in this middle range. This suggests that, despite comparable overall accuracy, the classical AI algorithmic DM exhibits more stable performance across distinct decision-maker profiles, while the LLM-based DMs struggle to generalize alignment in intermediate or ambiguous risk scenarios where trade-offs between cost and quality are less explicit.

As all three models demonstrated the highest alignment with the target having the lowest risk tolerance attribute value 0.0, followed by the high-risk target with attribute value of 1.0, these findings allow us to conjecture that both classical AI and LLM-based approaches may be learning from the data in similar ways, although they rely on fundamentally different mechanisms. The classical AI algorithmic DM relies explicitly on task-specific training data, whereas the LLM-based DM, despite not using any targeted fine-tuning, implicitly leverages its vast pretraining corpus and web-scale knowledge. This raises an important question: if we were to fine-tune or provide high-quality, domain-specific training data to the LLM, particularly for the moderate-risk target, which exhibited the lowest alignment, would its performance improve, or would it continue to depend primarily on its internal priors and pretraining biases? Future work will explore this direction through experiments that constrain the LLM-based DM to reason solely from structured, domain-specific inputs rather than its broader pretrained knowledge base.
\begin{table}[t]
\vskip -0.15in
\caption{Overall alignment accuracy for the three algorithmic DMs.}
\label{tab:overall-accuracy}
\begin{small}
\begin{center}
\begin{tabular}{l c}
\hline
\abovespace\belowspace
\textbf{Algorithmic DM} & \textbf{Overall Accuracy} \\
\hline

Classical AI Algorithmic DM & 0.892 \\
LRM-Based Algorithmic DM (GPT-5) & \textbf{0.895} \\
LLM-Based Algorithmic DM (GPT-4) & 0.892 \\

\hline
\end{tabular}
\end{center}
\vskip -0.10in
\end{small}
\end{table}

From a methodological perspective, both paradigms present distinct advantages and limitations. The key advantage of LLM-based algorithmic DMs lies in their accessibility, as the \textit{intelligent} reasoning component is already built in, allowing us to focus primarily on the experimental design and prompt formulation rather than developing a task-specific algorithm from scratch. However, this also introduces a critical limitation: the effectiveness of LLM-based DMs depends heavily on how well human language can represent nuanced cognitive constructs. For example, describing a moderate risk target (0.5) without overlapping linguistic cues from the extreme targets (0.0 and 1.0) is inherently difficult, raising the question of whether this limitation stems from the prompt itself or from the expressive boundaries of natural language. This limitation is particularly significant because representing all levels of cognitive attributes is essential to create alignable algorithmic DMs, since the literature in cognitive science consistently shows that human decision-making patterns, biases, and cognitive traits are highly nuanced and rarely confined to extremes. Capturing this middle ground is crucial for modeling realistic decision behavior and ensuring fine-grained alignment.

In contrast, classical AI algorithmic DMs allow alignment to be measured at a finer, more granular level. Rather than being restricted to three discrete targets, the classical DM can operationalize risk tolerance by defining targets at every 0.1 interval between 0 and 1, allowing us to study alignment at a finer level of granularity. This flexibility enables systematic exploration of the decision space, which LLMs, constrained by linguistic representation, would struggle to replicate. Additionally, variation in classical systems can be achieved through algorithmic modifications, while LLMs are largely limited to prompt tuning or model selection. The finding that both GPT-4 and GPT-5 achieved comparable alignment further suggests that this task may not require complex reasoning capabilities, as the non-reasoning model was successfully able to align the model's decisions with targets using weighted self-consistency.

\subsection{Challenges in Comparing Classical AI and LLM-Based Algorithmic DMs}
Comparing alignment performance between classical AI methods and LLM-based models presents inherent challenges. A key limitation lies in ensuring a fair comparison, as the type and amount of input data differ fundamentally across the two approaches. The LLM receives unstructured, natural language prompts with contextual cues, whereas the classical AI model operates on structured numerical features. This asymmetry means that even if both systems process equivalent decision scenarios, the nature of their input information and representational capacity is inherently different. Another challenge arises from variability in LLM outputs. LLMs exhibit stochasticity due to temperature and sampling parameters. To address this, we implemented a weighted self-consistency module, allowing the model to stabilize its decision tendencies across multiple generations while preserving context sensitivity.

Overall, these challenges highlight the trade-off between structure and adaptability in alignment research: classical AI algorithmic DMs offer structural stability and consistency, while LLM-based DMs provide contextual adaptability but require robust mechanisms to manage variability and ensure fair evaluation.


\section{Conclusions and Future Work}
In this work, we compared one classical AI and two LLM-based algorithmic decision-makers in a health insurance domain, investigating their ability to align to three targets with three levels of risk tolerance. Our findings reveal complementary strengths: LLM-based algorithmic DMs enable flexibility through prompt engineering to capture contextual cues for diverse targets, while classical approaches allow for more variations in granularity and maintain consistency across targets. Experimental results demonstrate that both classical AI and LLM-based DMs can be designed to align with different target decision-makers. Moreover, our successful reimplementation of the methodology proposed by Hu et al. on a different dataset further validates the robustness of their approach. 

While we closely followed Hu et al.’s methodology for the LLM-based implementation, several limitations remain. First, adapting the prompt structure to a new domain (health insurance) may introduce variability that affects comparability with prior work. Moreover, our study employs a different set of LLMs (GPT-4 and GPT-5), which differ in architecture and reasoning behavior from the models used in their work, potentially influencing alignment outcomes. Second, we did not vary parameters within the weighted self-consistency framework, which could reveal more nuanced relationships between prompt design and decision alignment. Additionally, our experiments relied on a static dataset rather than multiple iterative samples, which may limit the generalizability of the findings. Future work will address this by evaluating the average performance across multiple iterations. Moreover, while our experiment focused solely on risk tolerance, human decision-making is shaped by a rich interplay of cognitive attributes, such as ambiguity aversion, temporal discounting, and self-control, that exist on a continuum rather than at extremes. Capturing this full spectrum will be crucial for building alignable models in future studies.

For LLM-based algorithmic decision-makers, a key direction of investigation involves examining the limitations of natural language prompting itself. Specifically, we plan to explore prompting strategies that can represent nuanced cognitive attributes, such as moderate levels of risk tolerance, without relying on linguistic repetition or extreme descriptors. This raises a broader question about the inherent granularity and expressiveness of natural language as a medium for defining decision-making attributes.

\begin{acknowledgements} 
\noindent
This research was conducted as part of the In the Moment (ITM) project, supported by the Defense Advanced Research Projects Agency (DARPA) under contract number HR001122S0031. The views, opinions and/or findings expressed are those of the authors and should not be interpreted as representing the official views or policies of the Department of Defense or the U.S. Government. The authors also thank the reviewers for their insightful feedback, which helped improve the clarity and impact of this work.
\end{acknowledgements} 

\vskip -0.1in

{\parindent -10pt\leftskip 10pt\noindent
\bibliographystyle{cogsysapa}
\bibliography{format}
}

\end{document}